\documentclass{article}

\usepackage{arxiv}

\usepackage[utf8]{inputenc} 
\usepackage[T1]{fontenc}    
\usepackage{hyperref}       
\usepackage{url}            
\usepackage{booktabs}       
\usepackage{amsfonts}       
\usepackage{nicefrac}       
\usepackage{microtype}      
\usepackage{lipsum}
\usepackage{graphicx}
\usepackage{amsmath}  
\usepackage{amssymb}  
\usepackage{booktabs} 
\usepackage{array}    
\usepackage{tabularx}
\usepackage[normalem]{ulem}
\usepackage{tikz}
\usepackage{caption}  
\usepackage{hyperref}
\usepackage{multirow}
\usepackage{array}
\usepackage{makecell}
\usepackage{enumitem} 
\usepackage{fancyhdr}
\usepackage{subcaption}
\hypersetup{
colorlinks=true,
linkcolor=black
}

\graphicspath{ {./images/} }

\title{Software-Hardware Co-optimization for Modular E2E AV Paradigm: A Unified Framework of Optimization Approaches, Simulation Environment and Evaluation Metrics}

\author{
Chengzhi Ji$^{1}$,
Xingfeng Li$^{1}$,
Zhaodong Lv$^{1}$,
Hao Sun$^{2}$,
Pan Liu$^{1}$,
Hao Frank Yang$^{3}$,
\textbf{Ziyuan Pu$^{1,*}$} \\[0.5em]
$^{1}$School of Transportation, Southeast University, Nanjing 211189, China\\
$^{2}$School of Electronic Science and Engineering, Southeast University, Nanjing 211189, China\\
$^{3}$Department of Civil and Systems Engineering, Johns Hopkins University, Baltimore, MD 21218, USA\\
$^{*}$Corresponding author: \texttt{ziyuanpu@seu.edu.cn}
}

\begin{document}
\maketitle
\begin{abstract}
Modular end-to-end (ME2E) autonomous driving paradigms combine interpretability with global optimization capability and have achieved state-of-the-art performance. However, extant research has primarily emphasized improvements in accuracy metrics, while neglecting to address critical system-level considerations such as energy consumption and inference latency. Consequently, model designs have evolved to become increasingly intricate. To improve deployability, previous studies have investigated model compression and acceleration, yet these approaches are often pursued independently on either the software or hardware side. Software-only optimization cannot fundamentally eliminate intermediate tensor access and operator scheduling overheads. Hardware-only optimization is constrained by model structure and bit-width. Consequently, the benefits of such optimizations are often substantially diminished in real-world deployment. To address these limitations, this paper proposes a reusable software–hardware co-optimization and closed-loop evaluation framework for ME2E autonomous driving inference. The framework integrates software-level model optimizations with hardware-level computation optimizations under a unified system-level objective. Furthermore, a multidimensional evaluation metric, $\mathrm{EER}_{\mathrm{AV}}$, is introduced. This metric evaluates the ME2E autonomous driving system performance by jointly considering safety, comfort, efficiency, latency and energy, enabling quantitative assessment of the true system-level impact of different optimization strategies. The proposed framework is evaluated across multiple ME2E autonomous driving stacks. It preserves baseline-level accuracy while reducing inference latency by over 6× and per-frame energy to around one-fifth of the baseline. Furthermore, a 22.35\% improvement in the $\mathrm{EER}_{\mathrm{AV}}$ metric is achieved. These results validate that the proposed framework provides actionable optimization guidance from both software and hardware perspectives.
\end{abstract}

\keywords{Modular end-to-end \and Closed‑Loop Evaluation \and Software–Hardware co-optimization \and Energy Consumption}

\section{INTRODUCTION}

The modularized end-to-end (ME2E) paradigm employs a unified, differentiable training framework with shared features, addressing the complexities of integration, error accumulation, and the challenge of achieving global optimality in traditional modular autonomous driving systems \cite{1,hu2025multimodal}. Simultaneously, by maintaining modular decoupling, it enhances interpretability to some extent and mitigates the training difficulty associated with direct end-to-end approaches \cite{2,3}. 

The ME2E paradigm has demonstrated state-of-the-art performance in benchmarked driving tasks. However, its serial multi-task inference pipeline introduces substantial cumulative computational and latency overhead, which challenges in-vehicle edge platforms under tight compute and energy budgets. Although considerable progress has been made in the field of autonomous driving model compression and acceleration, many existing studies focus on localized, single-sided optimizations, either at the software or hardware level  \cite{33}. On the software side, network restructuring and model compression reduce nominal FLOPs, yet intermediate tensor traffic and operator scheduling overhead often remain, which limits the realized gains \cite{35}. On the hardware side, computation-graph optimization and memory management improve operator-level efficiency, but without co-optimizing the model architecture and numerical bit-width, the system still faces bandwidth demand and memory-capacity pressure \cite{36}. Consequently, single-sided optimization frequently leads to a mismatch between nominal computational reduction and realized system-level performance gains, and the benefits are often attenuated in deployment.

Beyond optimization, a fundamental challenge lies in system evaluation. The prevailing evaluation methodologies for autonomous driving models predominantly employ open-loop offline evaluation and closed-loop simulation-based testing paradigms \cite{18,20}. Both of these approaches rely on limited and highly structured scenarios with constrained data distributions, providing insufficient coverage of rare and safety-critical corner cases \cite{9,yang2025cross,hussain2024integrating}. As a result, they struggle to reflect model robustness and real-world operational behavior. Moreover, current evaluations primarily emphasize perception and decision accuracy, while largely overlooking deployment-critical metrics such as inference latency and energy consumption. However, system-level performance in autonomous driving is jointly determined by multiple factors, whose effects are non-linear and tightly coupled \cite{huang2024human, gao2023lane}. As a result, improvements achieved along a single dimension in simulation environments cannot be directly translated into practical performance gains on edge platforms \cite{zhong2023online,wu2024your}. More importantly, neither benchmark can quantitatively capture the impact of inference latency on safety, comfort, and efficiency. Consequently, efficient design remains an experience-driven trade-off rather than a process guided by clear and measurable optimization objectives.

These issues indicate that a significant gap remains between the algorithm-level design of ME2E autonomous driving techniques and their practical deployability in real-world scenarios. To address this gap, this paper proposes a co-optimization and evaluation framework tailored for the ME2E autonomous driving paradigm. The framework is motivated by the insight that software-side and hardware-side optimizations operate on different dimensions of deep learning models. Rather than being independent, these optimizations are intrinsically coupled at the system level and jointly influence downstream performance. Based on this observation, the main contributions of this work are summarized as follows:

\begin{enumerate}[label=\arabic*., leftmargin=*, itemsep=0pt, topsep=0pt]
\item A software--hardware co-optimization framework is proposed for the ME2E paradigm. Under a unified system-level objective, the framework jointly considers model compression techniques on the software side and operator acceleration methods on the hardware side, thereby broadening available optimization strategies for ME2E edge deployment.

\item A novel evaluation metric, $\mathrm{EER}_{\mathrm{AV}}$, is introduced. This metric characterizes model system performance across multiple dimensions, including safety, comfort, efficiency, latency, and energy consumption. It addresses a limitation of existing evaluation schemes that are primarily accuracy-oriented and fail to reflect practical deployability.

\item A real-time synchronous simulation method based on the CARLA Leaderboard is developed. This method enables quantitative analysis of how latency affects closed-loop performance metrics such as safety, comfort, and efficiency, thereby providing clear optimization targets for latency-aware model and system design.

\item The proposed method is systematically evaluated across multiple representative and advanced ME2E autonomous driving stacks, with extensive experiments conducted under a large number of corner-case simulation scenarios. The results demonstrate that the proposed framework significantly improves deployment-oriented metrics without compromising model accuracy. Specifically, system inference latency is reduced by more than $6\times$, per-frame energy consumption is reduced to around one-fifth of the baseline, and the composite evaluation metric $\mathrm{EER}_{\mathrm{AV}}$ is improved by up to $22.35\%$.
\end{enumerate}

This paper is outlined as follows: Section Two summarizes the existing lightweight design and evaluation methodologies pertaining to ME2E algorithms. Section Three delineates the three components of the evaluation framework: Software–Hardware co-optimization design space; a multi-dimensional evaluation approach and real-time synchronous simulation platform. Section Four presents the experimental design, results, and analysis. Section Five discusses the experimental results, highlighting key findings and open questions revealed by the proposed framework. Lastly, Section Six summarizes the contributions of this study and offers conclusions and recommendations for future research endeavors.
\section{LITERATURE REVIEW}
\label{sec:headings}

\subsection{End to End Autonomous Driving Systems} 

The emergence of end-to-end autonomous driving approaches aims to address several long-standing challenges in traditional modular autonomous driving systems, including complex system integration, error accumulation across modules, and the difficulty of achieving global optimization through separate training objectives. By jointly optimizing perception, prediction, and planning within a unified learning framework, end-to-end methods demonstrate improved system-level optimization capability and operational efficiency.

Recent studies have further validated and advanced the effectiveness of the end-to-end paradigm. For instance, Transfuser \cite{10,11} employs a multi-scale Transformer architecture to fuse image and LiDAR features. LAV \cite{12} leverages trajectories of all observed vehicles during training to alleviate the sparsity of end-to-end supervision signals. TCP \cite{13} integrates trajectory planning and control prediction into a dual-branch architecture to improve decision consistency. ThinkTwice \cite{14} combines future scene prediction with local feature retrieval to enable fine-grained perception of high-risk regions.

Despite these advances, direct end-to-end approaches still suffer from limited interpretability, difficulty in safety verification, and sparse training signals, which hinder their stable deployment and generalization in complex real-world scenarios. In contrast, the ME2E paradigm has attracted increasing attention by combining the global optimization capability of end-to-end training with the structural clarity of modular pipelines \cite{2,15}. This paradigm facilitates training and debugging while providing improved interpretability and verifiability. Representative works include P3 \cite{16}, which introduces a semantic occupancy grid as a unified intermediate representation across modules. Building upon this design, STP3 \cite{17} adopts a serial architecture to jointly perform perception, prediction, and planning in a module-wise manner. UniAD \cite{18} integrates full-stack driving tasks within a unified Transformer architecture, where intermediate representations from different modules collaboratively support final planning and decision-making. Although the ME2E paradigm achieves state-of-the-art performance, its serial multi-task architecture poses challenges in meeting real-time requirements, thereby limiting its practical deployment.

\subsection{Efficient End to End Autonomous Driving System Design for Edge Deployment}
Designing efficient autonomous driving algorithms remains challenging, largely due to the diverse architectural designs arising from different task decompositions \cite{shao2025hutformer}, combinations, and interaction mechanisms among subtasks. In response, a variety of optimization strategies have been proposed. VAD \cite{19,20} introduces a vectorized perception paradigm that enables parallel processing of map segmentation and object detection. SparseDrive \cite{21} integrates ego-vehicle trajectory planning with surrounding vehicle trajectory prediction into a synchronized process, facilitating joint planning and prediction. GenAD \cite{22} reformulates trajectory prediction as a generative problem, employing a VAE-GRU architecture to generate multimodal future trajectories in a latent space, thereby unifying prediction and planning. DriveTransformer \cite{23} adopts a task-level self-attention mechanism that enables direct interactions among multi-task queries and achieves full task parallelism.

In addition, the necessity of tightly coupled full-stack task integration has been questioned. Weng et al. conduct systematic ablation studies on UniAD and propose ParaDrive \cite{4}, exploring the feasibility of decoupling modules while maintaining strong performance in open-loop evaluation. However, due to the lack of closed-loop evaluation, the effectiveness of such designs in realistic driving scenarios remains insufficiently validated.

\subsection{Benchmark Evaluation System} 
Early open-loop evaluation paradigms, such as those adopted in nuScenes and Waymo, primarily rely on two metrics: L2 trajectory error and static collision rate. However, the validity of these metrics has been widely questioned \cite{7,8,9,24}. L2 trajectory error measures only the geometric discrepancy between predicted and ground-truth trajectories and is strongly influenced by vehicle speed. Meanwhile, collision rate metrics implicitly assume that surrounding traffic participants are unaffected by the ego vehicle’s behavior, thereby failing to assess interactive response capabilities. Furthermore, dataset imbalance further undermines evaluation effectiveness. For example, approximately 75\% of scenarios in nuScenes involve straight-line driving, with corner cases being underrepresented. As a result, even simple fully connected models, such as AD-MLP, can achieve high open-loop performance \cite{7}.

Closed-loop simulation partially addresses the lack of interaction modeling; however, existing platforms still exhibit notable limitations. Waymax \cite{25} and nuPlan \cite{26} adopt simplified geometric representations of vehicles and omit sensor characteristics such as noise, blind spots, and latency, which limits their suitability for end-to-end autonomous driving evaluation. Longest6 \cite{10}, an improved version of the CARLA Leaderboard V1, focuses primarily on basic driving skills and lacks support for complex multi-agent interactions. Although CARLA Leaderboard V2 \cite{27} introduces a driving score metric, its score decays rapidly during long-route evaluation, making it difficult to ensure stable and interpretable assessments. In addition, the absence of a standardized training dataset further hampers algorithm-level comparison \cite{28}.

Bench2Drive \cite{9} represents the latest benchmark designed to address these limitations. It includes 44 interaction scenarios and defines five advanced driving skill metrics, such as overtaking, lane merging, and emergency braking. Nevertheless, because Bench2Drive is built upon the CARLA Leaderboard architecture and adopts a synchronous simulation mechanism, it cannot realistically model inference and system response latency. This limitation restricts its ability to comprehensively evaluate the impact of real-time performance on autonomous driving behavior.

\section{METHODS}
\begin{figure}[!htbp]
  \centering
  \includegraphics[width=\linewidth]{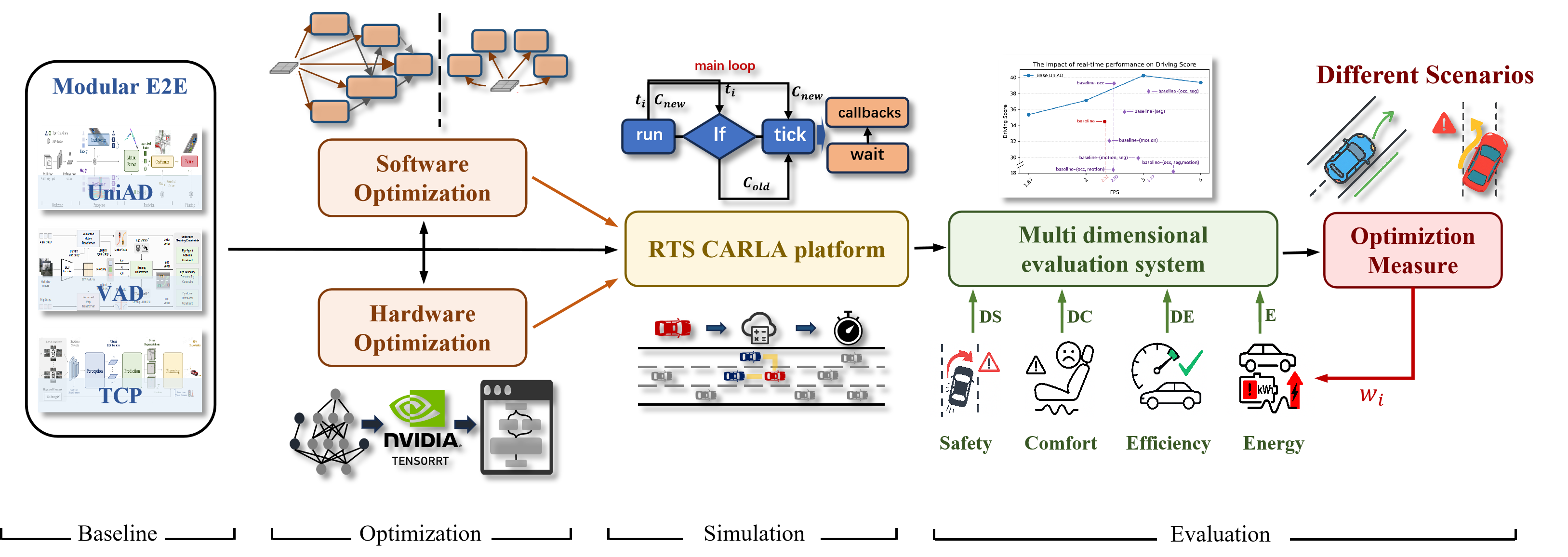}
  \caption{Overview of co-optimization and evaluation framework}
  \label{fig 1}
\end{figure}

This section introduces an optimization and evaluation framework designed to support both software- and hardware-level optimization for representative baseline ME2E autonomous driving models. The framework evaluates different optimization strategies across closed-loop driving performance and engineering deployability factors, and provides systematic guidance for selecting effective optimization configurations. \autoref{fig 1} illustrates the overall architecture of the proposed framework.

\subsection{Software–Hardware Co-optimization Framework}

This paper systematically explores the lightweight design space of ME2E autonomous driving algorithms through a software–hardware co-optimization framework. On the software side, the proposed framework considers two primary dimensions: structural decoupling and numerical precision compression, which are realized through module-wise pruning and module-wise quantization, respectively. On the hardware side, execution efficiency is improved through computation graph optimization and operator fusion, which reduce execution overhead and streamline the inference pipeline.

\subsubsection{Module-wise Pruning Design}
Full-stack ME2E autonomous driving models often exhibit functional overlap and computational redundancy across different modules. Improvements in perception or prediction accuracy provided by certain modules are sometimes marginal and may not offset the additional overhead imposed on real-time performance and computational resources. Systematic studies in ParaDrive \cite{4} demonstrate that, although joint multi-module training enhances feature sharing and global representation learning, planning decisions are not equally sensitive to the inference outputs of upstream modules. As a result, selectively ablating specific modules can improve inference efficiency while preserving closed-loop performance.

Building on this observation, this study extends ParaDrive’s module decoupling concept to a general ME2E computing framework and proposes a systematic optimization methodology for balancing module-level efficiency and accuracy contributions. The key design insight is that, in modular end-to-end architectures, the perception backbone provides a stable and information-rich representation for planning, whereas strict serial dependencies among downstream prediction modules are not always required. 

Based on this insight, the proposed framework realizes module-level separation and re-composition reasoning within a modular algorithmic structure by introducing direct information pathways between multiple prediction modules and the downstream planning module. This structural design decouples functional submodules, thereby enabling parallel and loosely coupled execution, while supporting independent evaluation and flexible reassembly. As a result, the respective impacts of individual components on closed-loop driving performance and resource utilization can be systematically assessed.

Using the UniAD framework as a case study, we restructure inter-module information flow to improve the accessibility of critical signals for planning while preserving the overall modular design. Specifically, direct connections from the map, tracking, and occupancy-related modules to the planning module are introduced, allowing the planner to directly consume their query representations without relying on multi-stage intermediate propagation. Meanwhile, core information pathways, such as BEV feature extraction, are retained to avoid degrading fundamental perception and decision-making capabilities.

\subsubsection{Module-wise Quantization Design}

As previously discussed, feature extraction modules in modular autonomous driving frameworks, such as the ResNet backbone, play a critical role in preserving perception accuracy and are therefore less amenable to aggressive pruning. At the same time, these modules account for a substantial portion of the inference-time computational cost. To balance efficiency and accuracy, this study adopts post-training quantization (PTQ) as a complementary optimization strategy, implemented in a module-wise manner across the network. 

PTQ facilitates efficient model compression without necessitating retraining. Prior studies have shown that PTQ can convert full-precision (float32) pre-trained networks into low-precision (int8) formats using only a small amount of calibration data, while maintaining near-original accuracy and significantly reducing computational cost. The process of quantizing a floating-point value $x$ into a low-precision integer approximation $q$ is shown in \eqref{1}.

\begin{equation}
    q=clip(round(\frac{x}{s})+z_{p},q_{min},q_{max})
\label{1}
\end{equation}
where $\operatorname{round}(\cdot)$ denotes the rounding-to-nearest operation, and $\operatorname{clip}(\cdot)$ constrains the results within the representable integer range. $q_{\min}$ and $q_{\max}$ represent the lower and upper bounds of the integer range, for example, from $-128$ to $127$ for int8. The dequantization process restores the integer tensor to its floating-point approximation, as expressed in \eqref{2}:

\begin{equation}
    \hat{x}=s\left(q-z_{p}\right)
\label{2}
\end{equation}
where $s$ denotes the quantization scale and $z_p$ denotes the zero point used to align the integer representation with zero. In symmetric quantization, $z_p$ is set to zero, and the scale $s$ is determined during calibration using a Max--Min calibration strategy, as defined in \eqref{3}.

\begin{equation}
    s=\frac{x_{max}-x_{min}}{q_{max}-q_{min}}
\label{3}
\end{equation}

The Max-Min quantization calibration strategy determines the quantization dynamic range by statistically calculating the maximum($x_{max}$) and minimum($x_{min}$) values of the tensor in the calibration data, aiming to sufficiently cover the feature distribution to reduce saturation error and maintain numerical stability in the quantization process across different modalities such as sparse point clouds and RGB features.

The calibration dataset consists of 256 frames randomly sampled from the Bench2Drive simulation dataset, ensuring that the collected activation distributions adequately reflect the model’s inference behavior under typical driving scenarios.

In order to achieve an equilibrium between the accuracy and performance of the quantized model, this study employs a two-stage node selection strategy during the quantization process. In the first stage, the Multi-Head Attention (MHA) structure is targeted, and key matrix multiplication nodes are filtered by means of tensor shape analysis. When the sequence length exceeds 512, the corresponding nodes are excluded from quantization to prevent excessive accumulation of quantization errors in the $QK^{T}$ dot product, which can be amplified during the Softmax computation under long-sequence conditions, leading to numerical instability of the attention distribution. This threshold aligns with the kernel support limitation reported in NVIDIA TensorRT documentation, where fused INT8 MHA operations are only validated for sequences up to 512 tokens \cite{37}.

In the second stage, the global computational graph is considered, and matrix multiplication nodes with degraded output dimensions (i.e., operations that degenerate from MatMul to GEMV) are further excluded. Since such operations exhibit low parallelism and cannot effectively utilize hardware acceleration units, forced quantization would introduce additional latency and performance degradation. Through this two-stage filtering mechanism, the proposed quantization process maintains numerical precision while improving efficiency and hardware adaptability.

\subsubsection{Hardware Computational Graph and Kernel Optimization Design}
\begin{figure}[htbp] 
    \centering
    \includegraphics[width=0.8\linewidth]{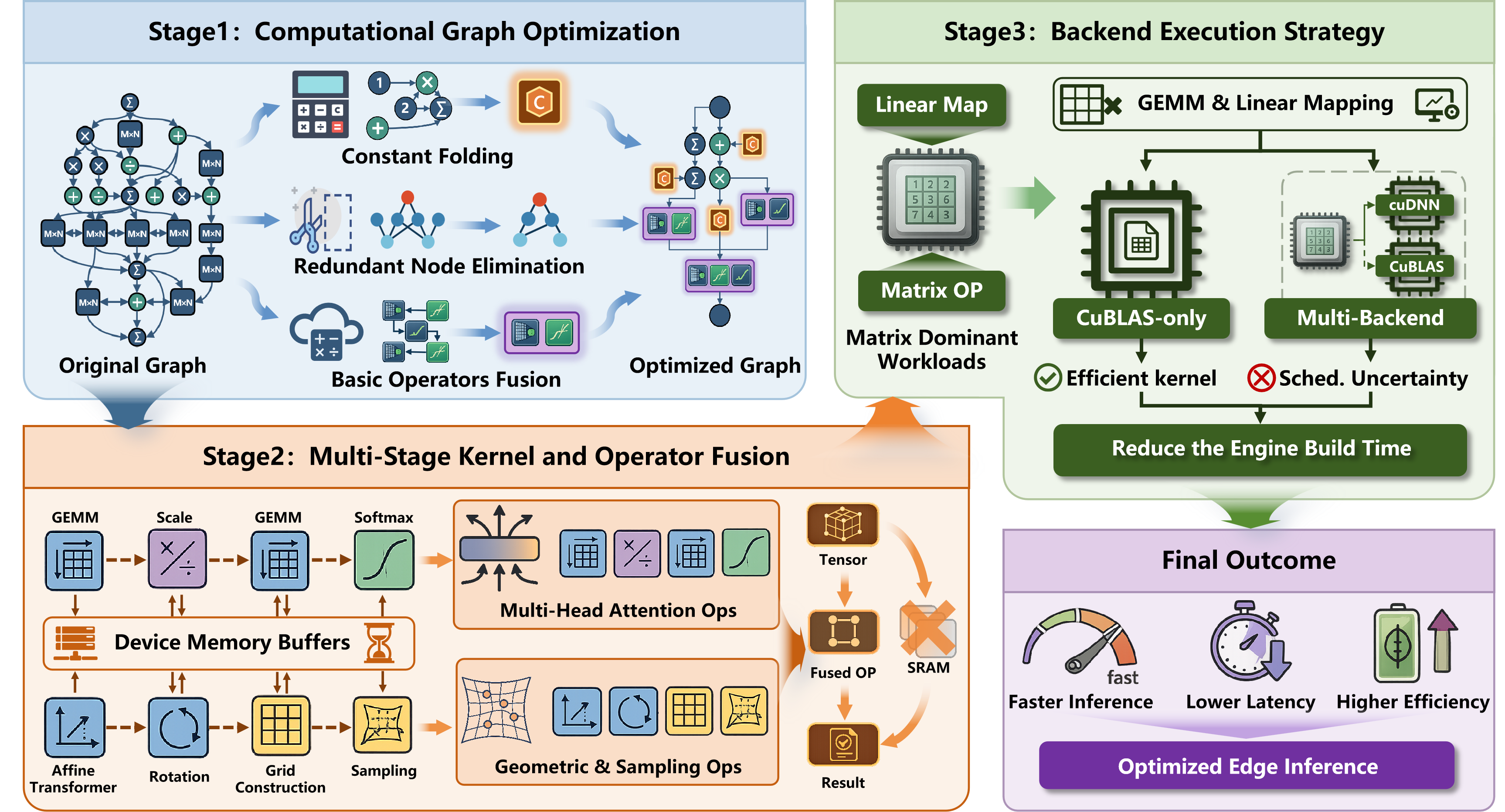}
    \caption{Multi-stage hardware optimization framework for computation graphs and kernels}
    \label{fig 3}
\end{figure}

To address the attenuation of algorithm-level optimization benefits under realistic deployment constraints, a hardware optimization method is proposed. This method performs both computational graph level and operator level optimizations, and is implemented based on TensorRT, as illustrated in \autoref{fig 3}.

At the computational graph level, computational efficiency is improved through graph rewriting using three complementary techniques. Constant folding precomputes operations involving fixed parameters, thereby eliminating redundant arithmetic during inference. Redundant node elimination removes unnecessary or duplicated nodes and branches, simplifying graph dependencies and reducing memory usage. Basic operator fusion combines adjacent low-level operators—such as convolution and addition—into a single composite operation, reducing runtime operator invocations and data transfer overhead. Collectively, these optimizations streamline the computational graph and improve inference efficiency.

Beyond graph-level rewriting, additional efficiency gains are achieved through operator level fusion targeting frequently executed multi-stage operators. 
For core computational modules such as Multi-Head Attention and Multi-Scale Deformable Attention, their internal multi-stage computations—including matrix mapping, GEMM, scaling, softmax, and weighted summation—are consolidated into unified kernels. 
In parallel, the same strategy is applied to geometry and sampling operators such as inverse, rotate, grid sampler and modulated deformable conv2d, where key internal computations are likewise integrated within one kernel. 
This reduces intermediate writebacks and kernel invocations, improving overall execution efficiency.

Furthermore, as the model computation is dominated by matrix multiplication and linear mapping with highly concentrated operator dependencies, only the cuBLAS backend is enabled during the tactic search stage. cuBLAS provides efficient matrix kernel implementations. Compared with simultaneously activating multiple backends such as cuDNN and cuBLASLt, this design choice reduces engine build time and mitigates scheduling uncertainties introduced by mixed library selection.

\subsection{Multidimensional Evaluation Framework for Driving Performance and Deployability}
This study evaluates the model in closed-loop simulation using a unified framework that jointly considers driving performance in terms of safety, efficiency, and comfort, as well as engineering deployability factors including inference latency and energy consumption. The proposed evaluation system is primarily realized through Real-Time Synchronous (RTS) simulation and the latency-Energy-Performance multi-dimensional evaluation approach.

\subsubsection{Real-time Synchronous Simulation Platform Design}
For the evaluation of the ME2E algorithm, most state-of-the-art studies primarily rely on the CARLA Leaderboard framework, which has employed a synchronous simulation mode since version~1.0. In this mode, the server waits until the client-side ego agent completes its computation and outputs the control command for the subsequent time step. The simulation world then advances by a fixed duration, referred to as the simulation time step ($\Delta t$), which represents the elapsed time between consecutive frames. Within this framework, $\Delta t$ can be adjusted to emulate the performance of an algorithm operating at an effective frequency of $1/\Delta t$ frames per second ($\mathrm{FPS}$) in real-world conditions.

Considering that the focus of this study is to explore the impact of inference latency on the effectiveness of closed-loop testing for algorithms, CARLA's default asynchronous mode could potentially serve this purpose. However, due to the absence of a unified clock control in this mode's simulation strategy, the server continuously outputs frames at high speed. This behavior may result in data loss or buffer overrun, potentially destabilizing the simulation process and leading to execution failures \cite{29}.

\begin{figure}[htbp] 
    \centering
    \includegraphics[width=\linewidth]{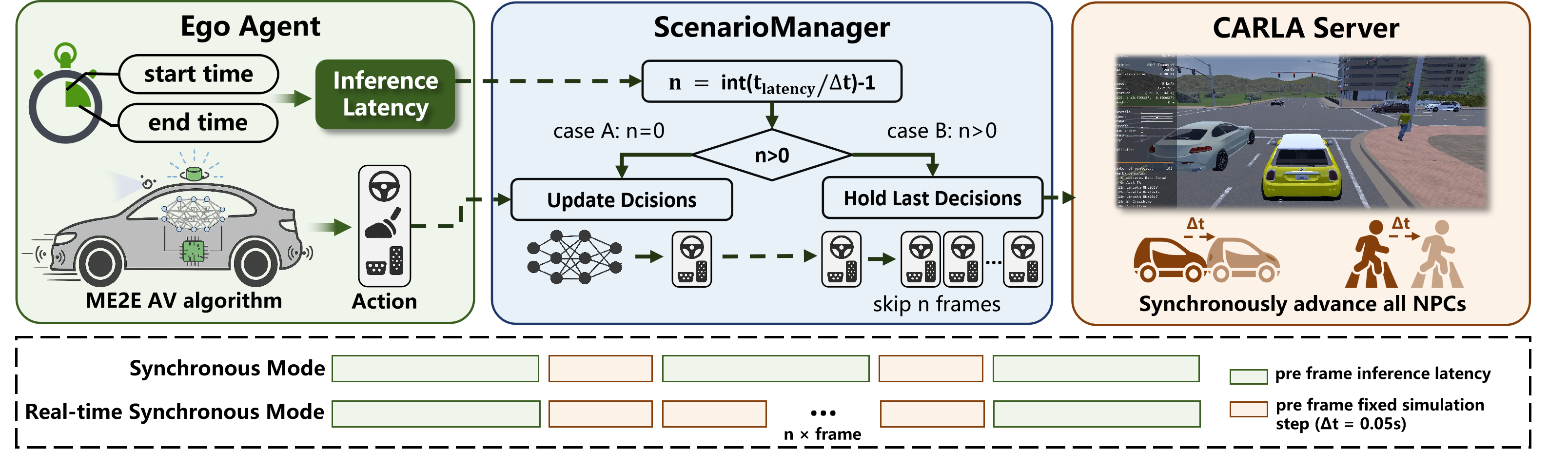}
    \caption{Implementation workflow of the RTS real-time synchronous simulation framework}
    \label{fig 4}
\end{figure}

To simulate the impact of inference latency on model performance while ensuring the stability of closed-loop simulation, this study proposes an optimization scheme to the conventional synchronous simulation framework (\autoref{fig 4}). Specifically, a time advancement strategy is introduced based on the actual inference latency: During each simulation frame, the system dynamically records the forward inference time of the algorithm and returns it to the \textit{ScenarioManager}, which then computes the corresponding number of simulation steps to advance according to Equation \eqref{4}. When $n>0$, this indicates that the inference latency exceeds one simulation step, and the client-side ego agent will skip $n$ decision updates and retain the previous control action (steering, throttle, braking) for the next $n+1$ frames, thereby modeling the effect of real-time inference latency on driving behavior.

\begin{equation}
    \mathrm{n} = \max\!\left(0,\; \mathrm{int}\!\left(\frac{t_i}{\Delta t}\right) - 1 \right)
\label{4}
\end{equation}
where $\Delta t$ represents the time increment of each update of CARLA simulation clock, which is generally 0.05 s by default. $t_{i}$ represents the reasoning latency, which is obtained by inserting a $time.perf\_counter$ timer at the start and end of the $run\_step$ function to achieve sub-second timing accuracy. For example, if $\Delta t=0.05 s$ and $t_{i}=0.13 s$ , then $n=\max\left(0, int(0.13/0.05)-1=1\right)$, meaning one control update is skipped and the ego maintains the previous action for two frames. Based on the tests conducted on the RTS simulation platform, the closed-loop performance of the algorithm considering the impact of latency can be measured.

\subsubsection{Composite Multidimensional Closed-loop Evaluation Metric}
To evaluate driving performance in closed-loop simulation, this study considers three complementary factors, driving safety (DS), driving efficiency (DE), and driving comfort (DC), which jointly characterize the overall quality of driving behavior.

DS follows the official CARLA evaluation protocol \cite{27} and measures the vehicle’s ability to complete route objectives while avoiding traffic violations. It is computed according to Eq.\eqref{5}:

\begin{equation}
    DS=\frac{1}{n}\sum_{i=1}^{n}RC_{i}*\prod_{j=1}^{n_{i}}P_{i,j}
\label{5}
\end{equation}
where $n$ denotes the number of road segments, $RC_{i}$ denotes the completion rate of the $i$-th road segment, and $P_{i,j}$ denotes the penalty coefficient for the $j$-th violation on the $i$-th road segment. DE reflects traffic efficiency by comparing the ego-vehicle speed with that of surrounding vehicles, while DC evaluates ride smoothness based on motion stability and dynamic constraints. The definitions of DE and DC are consistent with those used in Bench2Drive \cite{9}. They are computed as shown in Eqs.\eqref{6} and \eqref{7}, respectively:

\begin{equation}
    DE=\frac{1}{M}\sum_{k=1}^{M}\frac{v_{k}^{ego}}{\overline{v}_{k}^{near}}
\label{6}
\end{equation}

\begin{equation}
    DC=\frac{N_{smooth}}{N_{seg}}
\label{7}
\end{equation}
where DE is computed as the ratio between the ego-vehicle speed ($v_k^{ego}$) and the average speed of neighboring vehicles ($\bar{v}_k^{near}$) at a fixed number of speed checkpoints ($M=20$). During calculation, extreme outliers (greater than 1000\%) and trajectories that fail to pass the first 5\% of the road segment are removed to ensure a stable efficiency score. The DC metric is calculated by performing a Frame Variable Smoothness (FVS) threshold test on six dynamic variables, including longitudinal/lateral acceleration and angular velocity (\autoref{TABLE 1}), at the frame level. Then, a segment of 20 frames is evaluated, and only when the entire segment meets the FVS criteria is it counted as a “smooth segment”. The comfort level is ultimately measured by the proportion of smooth segments. If the trajectory remains blocked (speed $\leq$ 0.1~m/s for more than 60~s), it is treated as an acceptable stationary state, and no penalty is applied to DC. When the total number of frames in the trajectory is less than 20, the relevant route is excluded from DC statistics.

\begin{table}[htbp] \rmfamily
  \centering
  \caption{Expert threshold ranges for six dynamic variables}
  \label{TABLE 1}
  \begin{tabular}{ll} 
    \toprule
    Smoothness Variable & Expert threshold range \\
    \midrule
    longitudinal acceleration & \([-4.05,\ 2.40] \, \text{m/s}^2\) \\
    maximum absolute lateral acceleration & \([-4.89,\ 4.89] \, \text{m/s}^2\) \\
    yaw rate & \([-0.95,\ 0.95] \, \text{rad/s}\) \\
    yaw acceleration & \([-1.93,\ 1.93] \, \text{rad/s}^2\) \\
    longitudinal component of jerk & \([-4.13,\ 4.13] \, \text{m/s}^3\) \\
    maximum magnitude of jerk vector & \([-8.37,\ 8.37] \, \text{m/s}^3\) \\
    \bottomrule
  \end{tabular}
\end{table}

To demonstrate the impact of inference latency on performance, all results are obtained using the previously proposed synchronous real-time simulation framework. To distinguish these results from those reported under the standard CARLA Leaderboard evaluation setting, the three metrics are denoted as $\mathrm{DS}_{\mathrm{rt}}$, $\mathrm{DE}_{\mathrm{rt}}$, and $\mathrm{DC}_{\mathrm{rt}}$.

While measuring the inference speed and performance of the algorithm, this paper further explores the trade-off between energy consumption and closed-loop performance. GPU energy consumption data is collected via NVML (NVIDIA Management Library), and its Python wrapper library pynvml can be used to call \textit{nvmlDeviceGetTotalEnergyConsumption} to directly obtain the cumulative energy consumption (mJ) since the driver was loaded at the hardware level \cite{30,31,32}. 

Since the single-frame inference latency is reduced to below 0.1~s after TensorRT acceleration, instantaneous power fluctuations exceed the sampling resolution of \texttt{pynvml}. Therefore, a sliding-window method is adopted. Specifically, a statistical window of 100 frames is used to record the difference in cumulative energy consumption $E_{win}$ between the beginning and end of the window, and the average per-frame energy consumption $\bar{E}$ is computed as $E_{win}/100$. In addition, a 30-second warm-up period is applied before testing to mitigate frequency fluctuations during the initial GPU loading phase.

To comprehensively assess the benefits of different optimization strategies for autonomous driving algorithms after balancing performance and energy consumption, this study proposes the Autonomous Vehicle Energy Efficiency Ratio ($\mathrm{EER}_{\mathrm{AV}}$) metric. The metric is defined as shown in Eqs.~\eqref{8} and \eqref{9}.

\begin{equation}            
Q_{i}=w_{DS}x_{DS}+C(w_{DE}x_{DE}+w_{DC}x_{DC}+w_{E}x_{E});\quad C=
\begin{cases}
    1 & \mathrm{no~crash} \\
    0 & \mathrm{if~crash} 
    \end{cases}
\label{8}
\end{equation}

\begin{equation}
    \mathrm{EER}_{\mathrm{AV}}=\sum_{i=1}^{220}\mathrm{Q}_{i}/220
\label{9}
\end{equation}
where 220 corresponds to the total number of routes in the Bench2Drive dataset. $Q_i$ denotes the composite score for the $i$-th route. It represents a comprehensive system performance score that jointly accounts for safety, efficiency, comfort, latency and energy. The variables $x$ denote the normalized values of each indicator, and $w$ denote their corresponding weights. Note that $x_E$ represents normalized energy consumption (lower is better) and therefore enters Eq.~\eqref{8}  with a negative contribution. To reflect the human preference for absolute safety in driving decisions, we adopt a penalty design inspired by the nuPlan simulator \cite{26} and the CARLA leaderboard \cite{27}. A penalty multiplier $C$ is introduced to enforce substantially harsher deductions from the driving score in the event of a collision, thereby ensuring that safety maintains the highest priority within the overall evaluation framework.

To mitigate the influence of correlation among evaluation indicators, this study adopts the CRITIC method to determine the weights of the four indicators. The information content $c_i$ and the corresponding objective weight $w_j$ are computed as shown in Eqs.\eqref{10} and \eqref{11}.

\begin{equation}
    c_i=\sigma_i\sum_{j=1}^m\left(1-|r_{ij}|\right)
\label{10}
\end{equation}

\begin{equation}
    w_j=\frac{c_j}{\sum_{i=1}^{m}c_i}
\label{11}
\end{equation}
where $\sigma_{i}$ denotes the standard deviation of the $i$-th indicator, reflecting its contrast intensity, and $r_{ij}$ represents the correlation coefficient between the $i$-th and $j$-th indicators.

\section{RESULTS}

The experiments in this study consist of two parts: first, relying on the RTS simulation platform, the closed-loop performance of autonomous driving algorithms at different inference speeds is quantitatively evaluated by precisely controlling the single-inference latency of the agent; second, based on the proposed software–hardware co-optimization and evaluation framework, the baseline model is systematically optimized from both the software and hardware perspectives. Building upon this, power consumption monitoring and closed-loop simulation results are jointly incorporated, and different optimization schemes are uniformly evaluated using the accuracy–latency–energy multi-dimensional metric defined within the framework. Under the guidance of the framework, the overall autonomous driving optimization strategy with the best cost-effectiveness is ultimately identified.

This study uses the Bench2Drive dataset to build the CARLA simulation platform for closed-loop simulation. The dataset covers 44 representative interaction scenarios, including cutting-in maneuvers, lane changing and overtaking behaviors, as well as detouring and obstacle avoidance. Evaluations are conducted under diverse urban and weather conditions, ranging from sunny days in busy city centers to low visibility in dense fog. The evaluation protocol consists of 220 short routes, each approximately 150 meters in length.

This research experiment was conducted on a high-performance computing server equipped with 8 NVIDIA RTX 4090 GPUs and the NVIDIA driver version is 580.82.07, to support large-scale parallel inference and simulation computations. The software environment comprised the CARLA simulation platform (version 0.9.15), CUDA Toolkit 11.8, and the cuDNN 8.9.2 acceleration library. Hardware acceleration utilized the TensorRT 10.7 inference engine, with loaded ONNX models based on opset 13.

\subsection{Impact of Inference Latency on Driving Performance in Real-Time Simulation}
This paper aims to systematically explore the deployment-oriented optimization space for ME2E autonomous driving models, focusing on both software and hardware aspects. We select UniAD \cite{18} as the baseline because it is a representative ME2E framework and possesses state-of-the-art trajectory planning capability, serving as a reliable reference baseline.

Since the inference latency $t_i$ is inversely proportional to the frame rate $\mathrm{FPS}=1/ t_i$, changes in latency only result in a slow increase in $\mathrm{FPS}$ in the low frame rate range, but cause a sharp increase in $\mathrm{FPS}$ in the high frame rate range. For example, reducing the inference latency from 0.050 s to 0.033 s (a change of only 0.017 s) can increase the frame rate from 20 $\mathrm{FPS}$ to 30 $\mathrm{FPS}$. To balance resolution at both ends, this experiment employs a non-uniform sampling strategy: dense sampling in the low frame rate segment and gradually increasing the sampling interval in the high frame rate segment, ensuring more balanced distribution of sampling points along the $\mathrm{FPS}$ axis. This avoids redundant testing while accurately capturing critical performance change points.

\autoref{TABLE 2} shows the comprehensive impact of inference latency on key driving metrics. Overall, as inference latency decreases (i.e., $\mathrm{FPS}$ increases), the Driving Score shows a significant upward trend from 1 $\mathrm{FPS}$ to 20 $\mathrm{FPS}$, with the Driving Score improving by 20.33\%, indicating that both route completion rate and collision avoidance capability benefit from higher real-time feedback. When the frame rate reaches approximately 20-24 $\mathrm{FPS}$, performance improvements tend to plateau; further increasing inference speed not only yields limited marginal benefits but may also result in a slight decline. Through visual analysis of the vehicle's planned trajectory at the moment of the accident (\autoref{fig 6}), it can be observed that the UniAD algorithm lacks robustness in high-dynamic conflict scenarios, particularly when dealing with rapidly approaching targets, where trajectory estimation errors may occur, leading to modifications of the originally correct planned trajectory and resulting in collisions. This observation helps explain why overly high frame rates can degrade performance.
During the same process, comfort does not fluctuate significantly when $\mathrm{FPS}$ is less than 5, but it decreases rapidly as $\mathrm{FPS}$ continues to increase, dropping by 33.33\% at 20 $\mathrm{FPS}$. This phenomenon can be attributed to the agent adjusting driving strategies more frequently within shorter control cycles, resulting in less smooth driving. Driving efficiency remains largely constant, indicating that real-time performance is not the primary factor driving efficiency improvements.

Furthermore, based on the RTS simulation platform, we evaluated three models, including the tiny versions of UniAD and VAD, as well as TCP. Following the methodology described earlier, the frame drop count $n$ was computed during frame-by-frame inference to simulate inference performance under real-world conditions. To distinguish these results from the fixed inference latency experiments, all results obtained under the RTS setting are marked with superscripts. 

\begin{figure}[htbp]\rmfamily
  \centering

  \begin{minipage}[t]{0.53\textwidth}
    \centering
    \includegraphics[width=\linewidth]{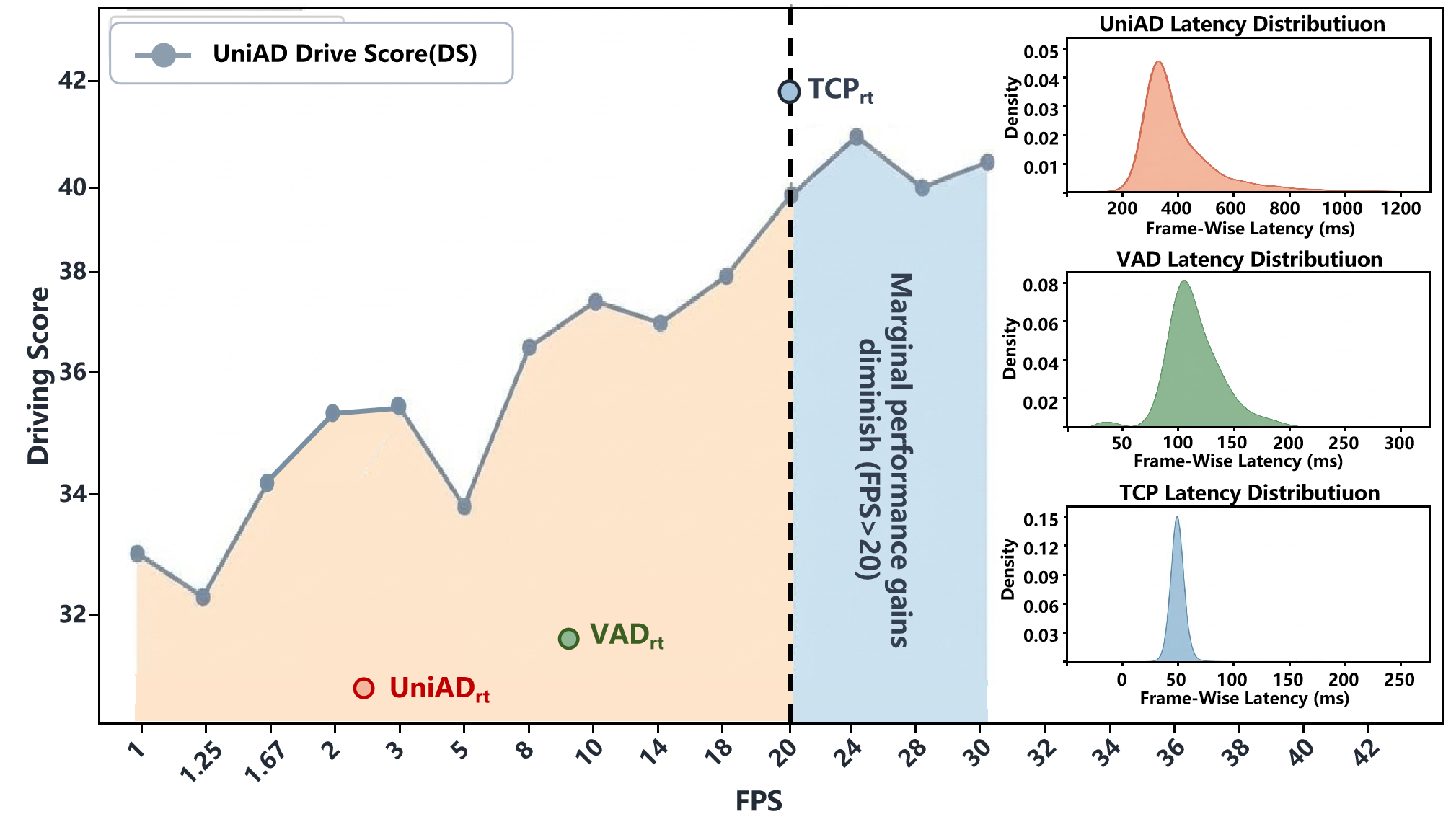}
    \vspace{3pt}
    \textbf{(a)} Safety
  \end{minipage}\hfill
  \begin{minipage}[t]{0.45\textwidth}
    \centering
    \includegraphics[width=\linewidth]{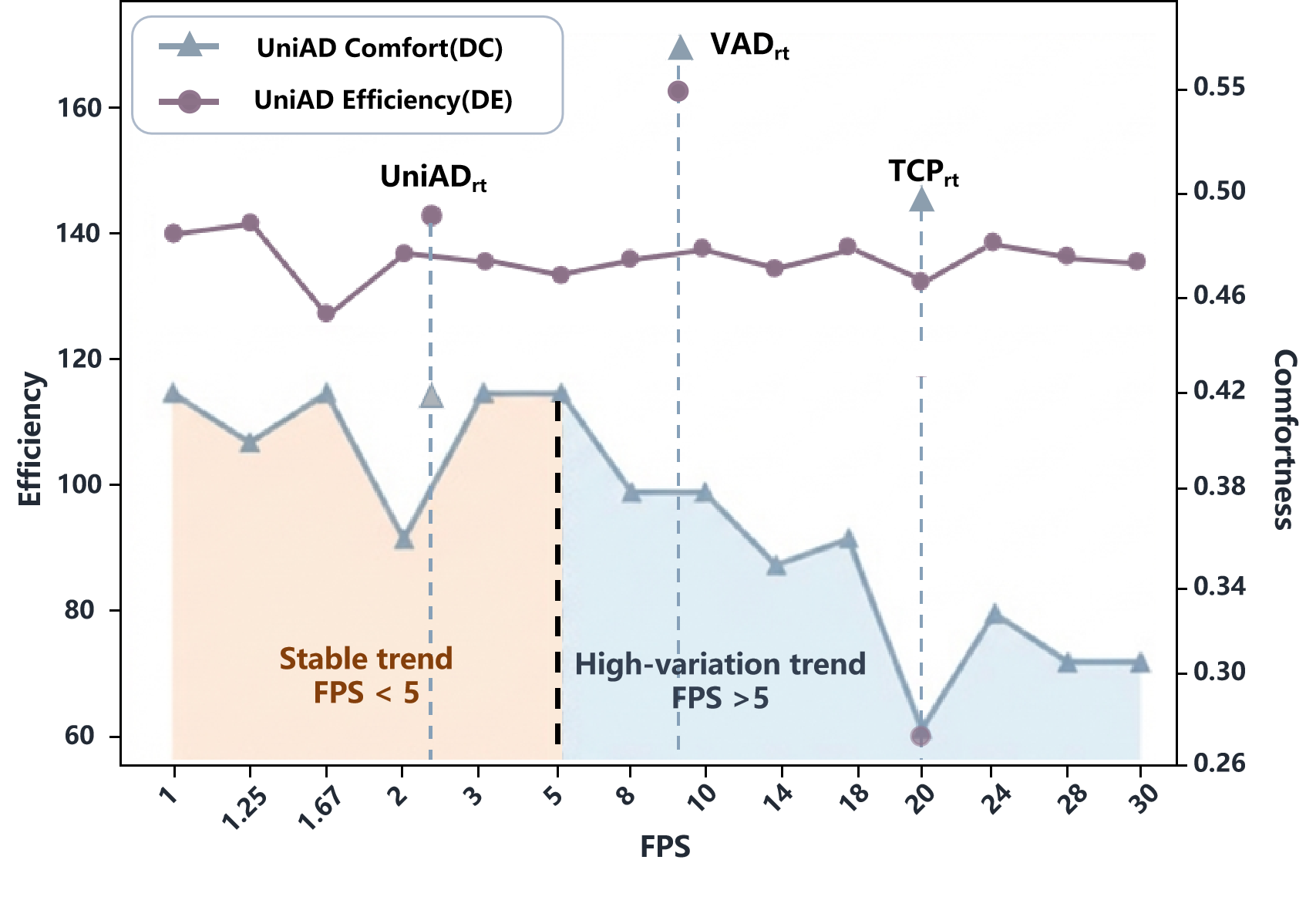}
    \vspace{3pt}
    \textbf{(b)} Efficiency and Comfort
  \end{minipage}

  \caption{The impact of real-time performance on closed-loop driving metrics}
  \label{fig 5}
\end{figure}

\begin{figure}[htbp] 
    \centering
    \includegraphics[width=0.8\linewidth]{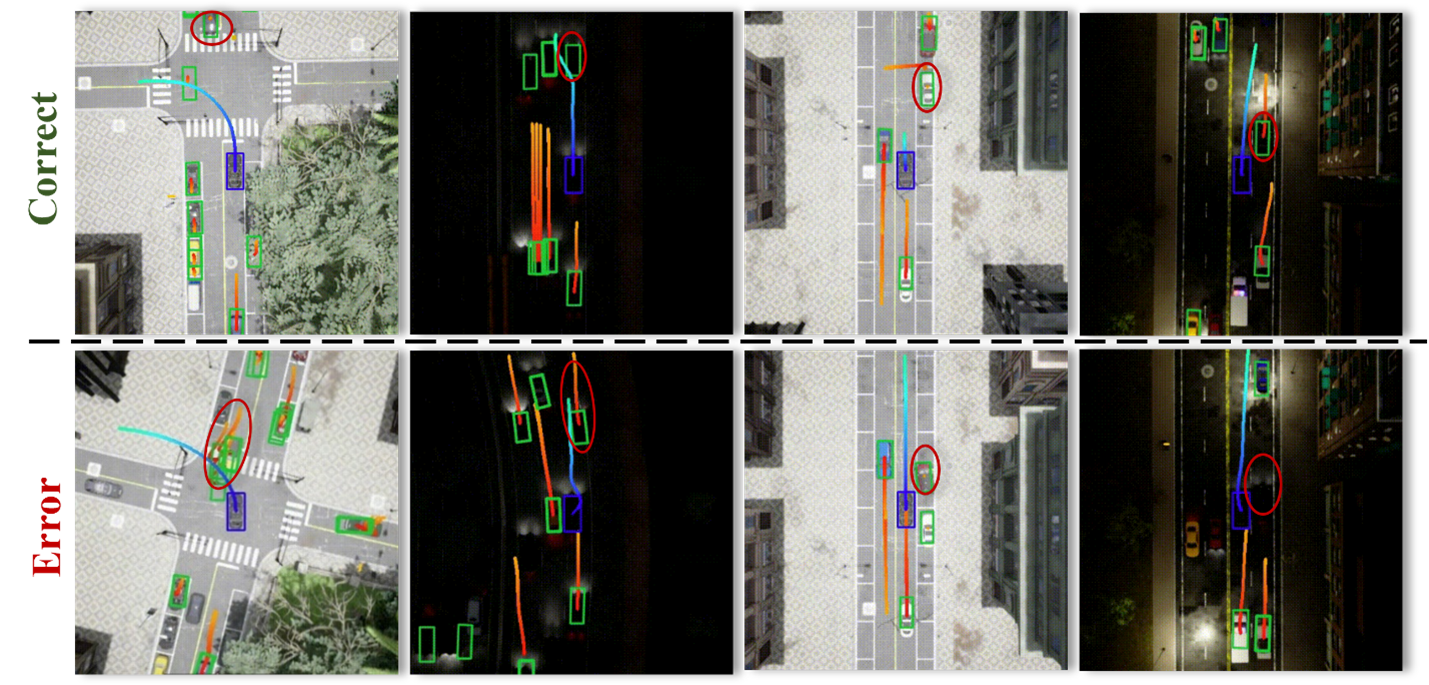}
    \caption{Visualization of vehicle trajectories at the accident moment, with the top showing a correct prediction and the bottom showing an incorrect prediction}
    \label{fig 6}
\end{figure}

The results in \autoref{fig 5} show that, under the same $\mathrm{FPS}$, UniAD$_{\mathrm{rt}}$ achieves a clearly lower Driving Score than UniAD with a fixed inference rate. VAD$_{\mathrm{rt}}$ also exhibits a substantial degradation relative to the Bench2Drive baseline (42.35), with a reduction of 24.62\%, whereas the performance of TCP$_{\mathrm{rt}}$ remains largely unchanged. Further analysis indicates that UniAD exhibits a pronounced long-tail inference latency distribution. A small fraction of frames incur significantly higher latency than the average, which leads to additional frame drops and degraded system-level performance. VAD exhibits a more concentrated per-frame latency distribution, however, it achieves significantly higher driving efficiency than UniAD (\autoref{fig 5}b), indicating a more aggressive and faster driving strategy that makes its closed-loop performance more sensitive to inference latency. Meanwhile, owing to TCP's relatively lightweight architecture, TCP is less affected by real-time constraints and is able to maintain stable inference performance under real-time simulation conditions.

In summary, the results demonstrate that inference latency has a substantial impact on closed-loop performance. Beyond average speed, we further find that multi-frame stability of inference latency is also a critical factor affecting overall closed-loop performance. The subsequent experiments will further validate and discuss this observation.

\begin{table}[htbp]\rmfamily
\centering
\caption{The relationship between ME2E algorithm driving performance and inference latency}
\label{TABLE 2}
\begin{tabularx}{\linewidth}{>{\centering\arraybackslash}m{1.8cm} *{4}{>{\centering\arraybackslash}X}}
\toprule
\textbf{Model} & \textbf{FPS} $\uparrow$ & \textbf{DS$_\mathrm{rt}$} $\uparrow$ & \textbf{DE$_\mathrm{rt}$} $\uparrow$ & \textbf{DC$_\mathrm{rt}$} $\uparrow$ \\
\midrule
\multirow{14}{*}{UniAD}
 & 1    & 32.85 & 139.81 & \textbf{0.42} \\
 & 1.25 & 32.04 & \textbf{141.38} & 0.40 \\
 & 1.67 & 34.17 & 126.89 & \textbf{0.42} \\
 & 2    & 33.50 & 136.69 & 0.36 \\
 & 3    & 35.59 & 135.45 & \textbf{0.42} \\
 & 5    & 33.74 & 133.40 & \textbf{0.42} \\
 & 8    & 36.70 & 135.57 & 0.38 \\
 & 10   & 37.55 & 137.44 & 0.38 \\
 & 14   & 37.15 & 134.28 & 0.35 \\
 & 18   & 38.01 & 137.55 & 0.36 \\
 & 20   & 39.53 & 132.03 & 0.28 \\
 & 24   & \textbf{40.63} & 138.21 & 0.33 \\
 & 28   & 39.68 & 136.24 & 0.31 \\
 & 30   & 40.15 & 135.90 & 0.31 \\
\midrule
UniAD$_\mathrm{rt}$ & 2.31 & 30.69 & 142.92 & 0.42 \\
VAD$_\mathrm{rt}$   & 9.52 & 31.92 & \textbf{166.26} & \textbf{0.54} \\
TCP$_\mathrm{rt}$   & 20   & \textbf{42.63} & 60.34 & 0.47 \\
\bottomrule
\end{tabularx}
\end{table}

\subsection{Impact of Module-Wise Pruning on Closed-Loop Performance}
The results in the previous section show that the RTS simulation platform can capture how changes in inference speed affect driving scores. In this section, we employ this platform to evaluate the effectiveness of the proposed optimization framework. To examine its generality, the strategy is applied to several representative ME2E autonomous driving models, including UniAD and VAD.\par
\begin{itemize}[leftmargin=*, itemsep=0pt, topsep=0pt]
\item UniAD features one of the most comprehensive sets of perception, prediction, and planning modules, providing a broad design space.
\item VAD adopts a similar network structure but incorporates a lightweight scene representation, achieving faster inference while maintaining core functionality.
\end{itemize}

\autoref{TABLE 3} presents the optimization results of all module-wise pruning design schemes, where (1-7) are for UniAD and (8-10) are for VAD.

Scheme~(1) shows that removing the OccFormer module alone results in a significant improvement in closed-loop performance, with the  $\mathrm{DS}_{\mathrm{rt}}$ increasing by 16.94\%. The performance of this configuration essentially returns to the baseline level reported in \autoref{TABLE 2}. This observation indicates that removing this module not only does not degrade closed-loop performance, but instead leads to a noticeable performance improvement. The underlying causes of this behavior will be analyzed in detail at the end of this section.

Ablating the Motion module leads to clear performance degradation in closed-loop evaluation. Four ablation schemes involving this module, namely schemes (3), (4), (5), and (7), obtain $\mathrm{DS}_{\mathrm{rt}}$ values that are substantially lower than those of the remaining schemes. These results indicate that predicting the future trajectories of surrounding agents is essential for generating appropriate interaction behaviors. Removing this capability reduces the model’s ability to anticipate potential conflicts during planning, which directly results in the observed degradation in closed-loop performance. Since open-loop metrics cannot fully capture interaction dynamics, these findings also help explain the discrepancy with ParaDrive’s open-loop evaluation results~\cite{4}.

For the semantic segmentation module, schemes~(2) and~(6) achieve slightly higher  $\mathrm{DS}_{\mathrm{rt}}$ values than the baseline, but at the cost of a notable increase in collision rate. Collisions with static structures account for 55.07\% of all collisions, far exceeding those observed in other schemes. This suggests that the segmentation module, by providing static semantic boundaries and drivable-area constraints, plays an important role in preventing intrusions into fixed obstacles.

The results of the VAD schemes~(8)--(10) are similar to those of UniAD. Pruning either the map or motion module leads to significant degradation in closed-loop performance. Owing to its relatively simpler structure, the performance degradation caused by such module pruning is even more severe, with reductions exceeding 33\% in most cases.

\begin{figure}[htbp] 
    \centering
    \includegraphics[width=\linewidth]{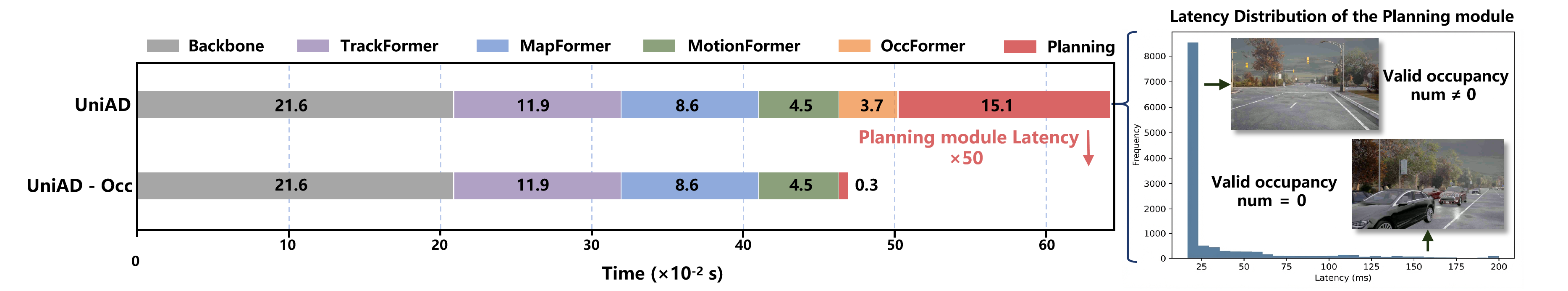}
    \caption{Module-wise inference latency breakdown and long-tail latency induced by OccFormer in UniAD}
    \label{fig 7}
\end{figure}

To further explain the performance gain from ablating OccFormer, we perform a more in-depth analysis. \autoref{fig 7} presents a detailed time consumption analysis of individual modules in UniAD. The planning module accounts for approximately 23\% of the total system inference latency, despite having a significantly simpler architecture compared to other modules. Frame-level latency analysis further reveals pronounced latency spikes in specific frames. These spikes primarily originate from the collision optimization submodule, which is activated only when the OccFormer module detects potential collision regions. This submodule employs a non-constrained optimization process with high computational complexity, resulting in per-invocation latency exceeding 150~ms. In real-world scenarios, this submodule is typically triggered when obstacles appear ahead of the ego vehicle, where stricter real-time constraints are imposed. However, under such conditions, the base version of UniAD introduces substantially longer inference latency. Consequently, the perception accuracy improvements provided by the OccFormer module are insufficient to compensate for the overall system performance degradation caused by increased inference latency in critical frames.

\begin{table}[htbp]\rmfamily
\centering
\caption{Performance evaluation of software optimization schemes based on module-wise pruning}
\label{TABLE 3}
\resizebox{\textwidth}{!}{
\begin{tabular}{l l c c c c c c}
\toprule
\textbf{Model} & \textbf{Scheme} &
\textbf{FPS $\uparrow$} &
\textbf{DS$_\mathrm{rt}$ $\uparrow$} &
\textbf{DE$_\mathrm{rt}$ $\uparrow$} &
\textbf{DC$_\mathrm{rt}$ $\uparrow$} &
\textbf{\makecell{Energy consumption \\ (J/frame) $\downarrow$}} &
\textbf{EER$_{AV}$ $\uparrow$} \\
\midrule

\multirow{8}{*}{UniAD}
& baseline & 2.31 & 30.69 & 142.92 & 0.43 & 191.29 & 23.31 \\
& (1) baseline - occ & \underline{2.59} & \textbf{35.89} & \underline{157.36} & \underline{0.40} & \underline{168.56} & \textbf{26.01} \\
& (2) baseline - seg & 2.72 & 32.24 & 150.43 & 0.42 & 175.22 & 24.03 \\
& (3) baseline - mot & 2.43 & 28.81 & 147.09 & 0.61 & 177.54 & 22.43 \\
& (4) baseline - (mot, seg) & 2.89 & 26.87 & 146.75 & 0.54 & 161.47 & 21.58 \\
& (5) baseline - (occ, mot) & 2.54 & 17.24 & 134.10 & 0.64 & 154.82 & 12.21 \\
& (6) baseline - (occ, seg) & 3.27 & 34.34 & 158.87 & 0.41 & 152.49 & 24.72 \\
& (7) baseline - (occ, seg, mot) & 3.58 & 16.80 & 136.92 & 0.64 & 138.74 & 9.85 \\

\midrule
\multirow{4}{*}{VAD}
& baseline & 9.52 & 31.92 & 166.26 & 0.54 & 31.88 & \textbf{26.11} \\
& (8) baseline - map & 10.56 & 21.53 & 148.96 & 0.45 & 31.22 & 18.84 \\
& (9) baseline - motion & 10.16 & 19.65 & 149.37 & 0.43 & 30.76 & 14.65 \\
& (10) baseline - all & 10.88 & 17.63 & 147.12 & 0.39 & 29.17 & 12.63 \\

\bottomrule
\end{tabular}
}
\end{table}

Overall, under the guidance of the proposed optimization framework, module-wise pruning improves system performance primarily by removing functional modules that exhibit pronounced inference-time fluctuations. However, experimental results indicate that its potential gains in inference speed and energy efficiency are limited. This limitation arises because the computational workload is largely concentrated in non-prunable components. For these critical computation modules, introducing quantization is therefore more appropriate. Accordingly, in the subsequent sections, the best-performing model from this section is adopted as a new baseline, and further optimizations are carried out along the optimization pathway defined by the proposed framework.

\subsection{Effect Analysis of Hardware Optimization and Quantization Strategies}
This section examines the impact of module-wise quantization and hardware optimizations, building on the module-wise pruning strategy described previously. It should be noted that the existing ME2E autonomous driving algorithms have relatively simple structures for each functional module except for the backbone and BEV encoder, and the number of Transformer layers is relatively small ($\leq 6$ layers). To maintain the acceleration effect while avoiding significant performance loss, INT8 quantization is applied to the model in this study. In addition, since quantizing a single module separately brings limited benefits and it is difficult to significantly demonstrate performance changes, the results are prone to noise interference and lack stability. Three quantization schemes are mainly designed:
\begin{enumerate}[label=\arabic*., leftmargin=*, itemsep=0pt, topsep=2pt] 
    \item Full network quantization;
    \item Quantization of the feature extraction module (backbone and BEV encoder);
    \item Quantization of the detection module (including other Functional modules).
\end{enumerate}

\autoref{TABLE 4} further lists the model sizes, node types, tensor ranges, and computationally intensive key nodes involved in each quantization scheme. 

\begin{table}[htbp]\rmfamily
\centering
\caption{Details of modular quantization schemes}
\label{TABLE 4}
\begin{tabular}{lcccc}
\toprule
\textbf{Quantization Scheme} & \textbf{{Tensors}} & \textbf{{Quantized Ops}} & \textbf{{Compute-intensive Ops}} & \textbf{{Model Size (MB)}} \\
\midrule
UniAD-Hardware\_op     & {-}   & {-}   & {-}   & 364.37 \\
Quant-Prediction    & 185   & 189   & 88    & 325.40 \\
Quant-Feature\_ext  & 365   & 281   & 138   & 257.17 \\
Quant-Full          & 543   & 460   & 220   & 241.59 \\
\midrule
VAD-Hardware\_op        & {-}   & {-}   & {-}   & 140.50 \\
Quant-Prediction     & 122   & 83    & 21    & 104.57 \\
Quant-Feature\_ext  & 72    & 72    & 56    & 119.25 \\
Quant-Full          & 194   & 155   & 77    & 83.32 \\
\bottomrule
\end{tabular}
\end{table}

\begin{table}[htbp]\rmfamily
\centering
\caption{Performance evaluation of module-wise quantization and hardware optimization}
\label{TABLE 5}
\begin{tabular}{lcccccc}
\toprule
\textbf{Scheme} & \textbf{FPS $\uparrow$} & \textbf{DS$_{\mathrm{rt}}$ $\uparrow$} & \textbf{DE$_{\mathrm{rt}}$ $\uparrow$} & \textbf{DC$_{\mathrm{rt}}$ $\uparrow$} & \begin{tabular}[c]{@{}c@{}}\textbf{Energy consumption}\\\textbf{(J/frame) $\downarrow$}\end{tabular} & \textbf{EER$_{\text{AV}}$ $\uparrow$} \\ \midrule
UniAD-Hardware\_op  & 22    & 38.92 & 161.64 & 0.29 & 27.89 & 28.13 \\
Quant-Prediction  & 25    & 29.53 & 143.91 & 0.32 & 23.36 & 22.25 \\
Quant-Feature\_ext & \underline{28.3}  & \underline{36.59} & \underline{152.35} & \textbf{0.39} & \underline{20.71} & \textbf{28.52} \\
Quant-Full       & 31    & 29.57 & 149.74 & 0.35 & 18.64 & 22.65 \\ \midrule
VAD-Hardware\_op    & 57.80 & 34.46 & 158.36 & 0.59 & 6.89  & 26.32 \\
Quant-Prediction  & 66.67 & 29.83 & 160.33 & 0.52 & 6.14  & 22.54 \\
Quant-Feature\_ext & \underline{69.93} & \textbf{34.95} & \underline{156.75} & \textbf{0.63} & \underline{4.27} & \textbf{26.44} \\
Quant-Full       & 71.43 & 28.80 & 160.91 & 0.50 & 3.86  & 22.74 \\ \bottomrule
\end{tabular}
\end{table}

\autoref{TABLE 5} presents the overall performance of the models under the optimization of the proposed module-wise quantization and hardware acceleration schemes. The results show that the hardware acceleration scheme brings significant improvements to both autonomous driving models, achieving inference speedups of $8.49\times$ for UniAD and $6.07\times$ for VAD, while the closed-loop performance $\mathrm{DS}_{\mathrm{rt}}$ is improved by 8.44\% and 7.95\%, respectively.

In addition, the table further reports a performance comparison of three different quantization schemes supported by the proposed system. It can be observed that quantization methods can effectively increase model throughput. Full-network quantization yields inference speed improvements of 41\% for UniAD and 24\% for VAD, accompanied by $\mathrm{DS}_{\mathrm{rt}}$ drops of 24.02\% and 16.42\%, respectively. Notably, compared with other quantization schemes, quantizing the feature-extraction module provides the largest acceleration benefit. The resulting impact on overall driving performance is limited and remains within an acceptable range, while also achieving the best performance in terms of $\mathrm{EER}_{\mathrm{AV}}$. These results suggest that, in scenarios requiring a balance between real-time performance and overall driving performance, prioritizing low-bit quantization for the perception backbone while adopting mixed precision or maintaining high precision for the planning branch constitutes a more cost-effective engineering trade-off strategy.

To further examine whether extremely high frame rates are necessary, this study reduces the model’s throughput by inserting time.sleep during inference. As shown in \autoref{TABLE 6}, decreasing the frame rate from 22 $\mathrm{FPS}$ to 10 $\mathrm{FPS}$ leads to only a 7.15\% drop in performance. Since the energy consumption per frame remains roughly constant, the average power satisfies $P=E_{frame} \times \mathrm{FPS}$ and thus decreases approximately linearly with frame rate. Compared with the original setting, the overall power consumption is reduced by about 50\%. These results indicate that, without modifying the model architecture, moderately lowering the frame rate can substantially reduce energy and power costs while preserving closed-loop performance, offering an effective runtime energy–efficiency trade-off for resource-constrained platforms.

\begin{table}[htbp]\rmfamily
\centering
\caption{Impact of Inference Latency on driving performance and Power Consumption}
\label{TABLE 6}
\begin{tabular}{lccccc}
\toprule
\textbf{Scheme} & \textbf{FPS $\uparrow$} & \textbf{DS$_{\mathrm{rt}}$ $\uparrow$} & \textbf{DE$_{\mathrm{rt}}$ $\uparrow$} & \textbf{DC$_{\mathrm{rt}}$ $\uparrow$} & \textbf{Power (W) $\downarrow$} \\
\midrule
\multirow{3}{*}{UniAD-Hardware\_op }
 & 22 & 38.92 & 161.64 & 0.29 & 613.58 \\
 & 10 & 36.13 & 141.82 & 0.32 & 246.21 \\
 & 4  & 29.95 & 140.35 & 0.38 & 138.56 \\
\bottomrule
\end{tabular}
\end{table}

\section{DISCUSSION}
This study focuses on ME2E autonomous driving systems and conducts extensive experiments based on the proposed software–hardware co-optimization and multidimensional evaluation framework. The results reveal a set of system-level patterns that are difficult to observe using conventional single-dimension evaluation metrics. These findings partially complement and revise existing understandings of the ME2E paradigm, particularly regarding the relationships among accuracy, latency, and energy consumption across multiple performance dimensions.

\subsection{Summary of Main Findings}
\textbf{High frame rate is not optimal, and latency optimization requires explicit objectives}: The results indicate that the relationship between inference speed and closed-loop driving performance is nonlinear. In low frame-rate regimes, reducing inference latency significantly improves the Driving Score. When the frame rate exceeds approximately 10 $\mathrm{FPS}$, the performance gain diminishes and peaks around 20–24 $\mathrm{FPS}$. Further acceleration may even degrade performance. This behavior indicates that excessively high control update frequencies can amplify perception noise and short-horizon prediction errors, thereby reducing safety and comfort. From an energy-efficiency perspective, high frame rates are also suboptimal. Since per-frame energy consumption remains nearly constant, average system power increases approximately linearly with inference frame rate, while performance gains saturate at high $\mathrm{FPS}$. As a result, moderately reducing $\mathrm{FPS}$ causes only limited degradation in closed-loop performance but can substantially reduce overall power consumption (see \autoref{TABLE 6}). These observations indicate that latency optimization should be guided by clearly defined objectives.

\textbf{Multi-frame stability of inference latency is critical}: The experiments demonstrate that latency variability has a significant impact on system performance. Even when average $\mathrm{FPS}$ values are comparable, models with long-tailed latency distributions toward high-latency events exhibit noticeably lower Driving Scores. As shown in \autoref{TABLE 2}, UniAD$_{\mathrm{rt}}$ exhibits substantially degraded performance. This degradation arises because the conditionally triggered collision optimization module in UniAD introduces large latency spikes in a small number of critical frames. These spikes delay control updates and increase collision risk. This result shows that evaluating latency solely based on average inference speed is insufficient to characterize system safety, as real driving performance is often dominated by inference latency in a small number of critical frames.

\textbf{Submodules contribute highly unevenly to closed-loop performance in the ME2E paradigm}: Module ablation experiments indicate that different functional submodules in the ME2E paradigm contribute unevenly to system performance. Modules closely coupled with environment interaction play a decisive role, such as behavior prediction. Removing such modules significantly degrades Driving Score. In contrast, some modules designed primarily for safety enhancement may not yield corresponding benefits under latency constraints. In some cases, they even harm system stability by introducing additional latency variability. These results demonstrate that module design and evaluation based solely on open-loop accuracy are insufficient. Multidimensional closed-loop evaluation is essential to comprehensively balance system-level performance.

\textbf{Software–hardware co-optimization is a key pathway toward system-level optimality}: The experiments show that the main computational bottleneck of the ME2E paradigm lies in the perception feature extraction stage, while the planning and decision branches are more sensitive to numerical precision. This structural heterogeneity makes single-dimension optimization inadequate for achieving system-level optimality. Software-side optimization can reduce model complexity, but its latency and energy benefits are limited by the non-prunability of critical functional modules. Hardware-side computation graph optimization and operator fusion can significantly improve execution efficiency. However, without coordinated analysis of model structure, full-network optimization may lead to substantial performance degradation. These observations suggest that coordinated software–hardware co-optimization offers a more effective pathway toward system-level optimization under the ME2E paradigm.

\subsection{Limitations and Future Directions}
Although the proposed framework demonstrates strong optimization capability, several limitations remain. The current study focuses primarily on ME2E autonomous driving systems, and its applicability to more general intelligent system paradigms has yet to be explored. Moreover, experimental results indicate that different models exhibit distinct driving strategies, leading to variations in driving efficiency and, consequently, different sensitivities to inference latency; however, the specific architectural or algorithmic design factors responsible for these differences have not been systematically investigated. In addition, the impact of inference latency appears to be highly scenario-dependent, with interaction-intensive scenarios such as lane changing being more sensitive to latency than steady-state driving, yet a quantitative characterization of latency–performance relationships across diverse scenarios is still lacking. These observations provide important insights for future research. Future work may extend this study to multi-agent scenarios, such as vehicle–infrastructure cooperation and air–ground coordination, to further validate the effectiveness and generality of the proposed approach. In addition, more systematic and in-depth investigations will be conducted to explore the intrinsic relationships among model design, scenario characteristics, and latency requirements.

\section{CONCLUSIONS}
In this work, we propose a software–hardware co-optimization and closed-loop evaluation framework for ME2E autonomous driving systems. The framework aims to address the limitations of prior studies that focus on single-sided software or hardware optimization in practical deployment, as well as the inability of existing evaluation schemes with limited dimensions to reflect true deployment performance. The framework jointly optimizes software and hardware components. It reduces system latency by more than 6× while preserving baseline model accuracy, lowers per-frame energy consumption to around one-fifth of the baseline, and achieves up to a 22.35\% improvement in the composite $\mathrm{EER}_{\mathrm{AV}}$ metric. Further analysis of the experimental results reveals a set of system-level patterns. These results further demonstrate the necessity of software–hardware co-optimization and multidimensional joint evaluation.

Overall, this study provides new insights and tools for the efficient design and deployment-oriented evaluation of end-to-end autonomous driving algorithms. Future work will validate the proposed approach on real-vehicle systems and explore general acceleration frameworks for broader classes of ME2E intelligent systems.

\bibliographystyle{unsrt}  
\bibliography{references}  

\end{document}